\documentclass[letterpaper, 10 pt, conference]{ieeeconf}  % Comment this line out if you need a4paper

\IEEEoverridecommandlockouts                              % This command is only needed if 
\usepackage[T1]{fontenc}
\usepackage{graphicx}
\usepackage{acro}
\usepackage{xcolor}
\usepackage{subcaption}
\usepackage[switch]{lineno}
\usepackage{multirow}
\usepackage{multicol}
\usepackage{booktabs}
\usepackage{amssymb}
\usepackage{bm}
\usepackage{algorithm}
\usepackage{algorithmicx}
\usepackage{algpseudocode}
\usepackage{soul}
\usepackage{amsmath} 
\usepackage{dsfont}
\usepackage{hyperref}

\DeclareAcronym{DOF}{
short=DoF,
long=degrees of freedom}

\DeclareAcronym{DH}{
short=DH,
long=Denavit-Hartenberg}

\DeclareAcronym{BFGS}{
short= BFGS,
long= Broyden–Fletcher–Goldfarb–Shanno}

\DeclareAcronym{L-BFGS}{
short= L-BFGS,
long= Limited-memory Broyden–Fletcher–Goldfarb–Shanno}

\DeclareAcronym{L-BFGS-B}{
short= L-BFGS-B,
long= Limited-memory Broyden–Fletcher– Goldfarb–Shanno-Bounded}

\DeclareAcronym{FK}{
short=FK,
long=forward kinematics}

\DeclareAcronym{IK}{
short=IK,
long=inverse kinematics}

\DeclareAcronym{GPU}{
short=GPU,
long=Graphics Processing Unit}

\DeclareAcronym{PDOMP}{
short=PDOMP,
long=Propagative Distance Optimization for Motion Planning}

\DeclareAcronym{PDO-IK}{
short=PDO-IK,
long=Propagative Distance optimization for Constrained Inverse Kinematics}

\DeclareAcronym{DP}{
short=DP,
long=dynamic programming}

\DeclareAcronym{SQP}{
short=SQP,
long=sequential quadratic programming}

\DeclareAcronym{CoM}{
short=CoM,
long=center of mass}

\DeclareAcronym{PCA}{
short=PCA,
long=Principal Component Analysis}

\title{\LARGE \bf
Adversarial Game-Theoretic Algorithm for Dexterous Grasp Synthesis
}

\author{Yu Chen$^{1}$,
Botao He$^{2}$,
Yuemin Mao$^{1}$,
Arthur Jakobsson$^{1}$, 
Jeffrey Ke$^{1}$, \\
Yiannis Aloimonos$^{2}$, 
Guanya Shi$^{1}$,
Howie Choset$^{1}$,
Jiayuan Mao$^{3}$, and 
Jeffrey Ichnowski$^{1}$
\thanks{$^{1}$Robotics Institute, Carnegie Mellon University, PA 15213. 
$^{2}$Department of Computer Science, University of Maryland, MD 20742.
$^{3}$Department of Computer and Information Science, University of Pennsylvania, PA 19104.}
\thanks{Email: \texttt{yuchen3@andrew.cmu.edu}.}
}

\begin{document}

\maketitle
\thispagestyle{empty}
\pagestyle{empty}

%%%%%%%%%%%%%%%%%%%%%%%%%%%%%%%%%%%%%%%%%%%%%%%%%%%%%%%%%%%%%%%%%%%%%%%%%%%%%%%%
\begin{abstract}
For many complex tasks, multi-finger robot hands are poised to revolutionize how we interact with the world, but reliably grasping objects remains a significant challenge.
We focus on the problem of synthesizing grasps for multi-finger robot hands that, given a target object's geometry and pose, computes a hand configuration. 
Existing approaches often struggle to produce reliable grasps that sufficiently constrain object motion, leading to instability under disturbances and failed grasps.
A key reason is that during grasp generation, they typically focus on resisting a single wrench, while ignoring the object's potential for adversarial movements, such as escaping.
We propose a new grasp-synthesis approach that explicitly captures and leverages the adversarial object motion in grasp generation by formulating the problem as a two-player game. %, in which the players each solve their own optimization problem. 
One player controls the robot to generate feasible grasp configurations, while the other adversarially controls the object to seek motions that attempt to escape from the grasp.
Simulation experiments on various robot platforms and target objects show that our approach achieves a success rate of 75.78$\%$, up to 19.61$\%$ higher than the state-of-the-art baseline. 
% The two-player game mechanism improves the grasping success rate by 27.40$\%$ over the method that optimizes only the robot configuration. 
The two-player game mechanism improves the grasping success rate by 27.40$\%$ over the method without the game formulation. 
Our approach requires only 0.28--1.04 seconds on average to generate a grasp configuration, depending on the robot platform, making it suitable for real-world deployment.
In real-world experiments, our approach achieves an average success rate of 85.0$\%$ on ShadowHand and 87.5$\%$ on LeapHand, which confirms its feasibility and effectiveness in real robot setups.
Code is publicly available at \texttt{https://github.com/Neuling-jpg/Game4Grasp}.
\end{abstract}

%%%%%%%%%%%%%%%%%%%%%%%%%%%%%%%%%%%%%%%%%%%%%%%%%%%%%%%%%%%%%%%%%%%%%%%%%%%%%%%%
\section{Introduction}
The world is filled with objects and tools explicitly designed for human hands. As a new generation of dexterous robot hands becomes a reality, we must equip them with the ability to navigate this human-centric environment by mastering the core task of grasping.
Grasp synthesis is a fundamental problem in robot manipulation that seeks a grasp configuration of the articulated robot hand given the pose and geometry of the target object. 
For low-dimensional robot hands, such as two- or three-finger grippers, this can be solved through sampling- or search-based strategies~\cite{miller2004graspit,gualtieri2016high,liu2021synthesizing} in seconds. 
For more complicated robots, sampling-based approaches often take several minutes to find a feasible grasp~\cite{wei2024d}. 
For the purpose of online grasp generation, more recent approaches leverage optimization-~\cite{dai2017synthesis} or learning-based~\cite{wei2024d} techniques for efficient grasp synthesis. 

% However, the resulting grasps of these methods often fail to sufficiently constrain the object motion. 
However, the resulting grasps of these methods often fail to sufficiently constrain the object from escaping,
which often results in unstable or failed grasps, particularly under external disturbances or during tool using.
% To address this challenge, we propose a new grasp synthesis approach that generates reliable grasps within seconds.
% Our approach is based on the key insight that most existing methods typically ignore the potential adversarial motions of the object during the generation of the grasp configuration. 
This is because most existing optimization- or learning-based methods typically focus on resisting a wrench, such as gravity, and ignore potential adversarial motions of the object.
%during the generation of the grasp configuration. 
Although some work introduce adversarial techniques to grasp synthesis~\cite{wang2025advgrasp,wang2019adversarial,jing2021domain}, they are usually employed as post hoc evaluation or filtering strategies rather than being explicitly incorporated into the grasp generation process itself. 
% \jiayuan{usually resulting in what?}
Consequently, the generated grasps may appear feasible under static assumptions but often lack robustness in constraining the pose of the object under perturbations.

\begin{figure}[t]
    \centering
    \includegraphics[width=0.95\linewidth]{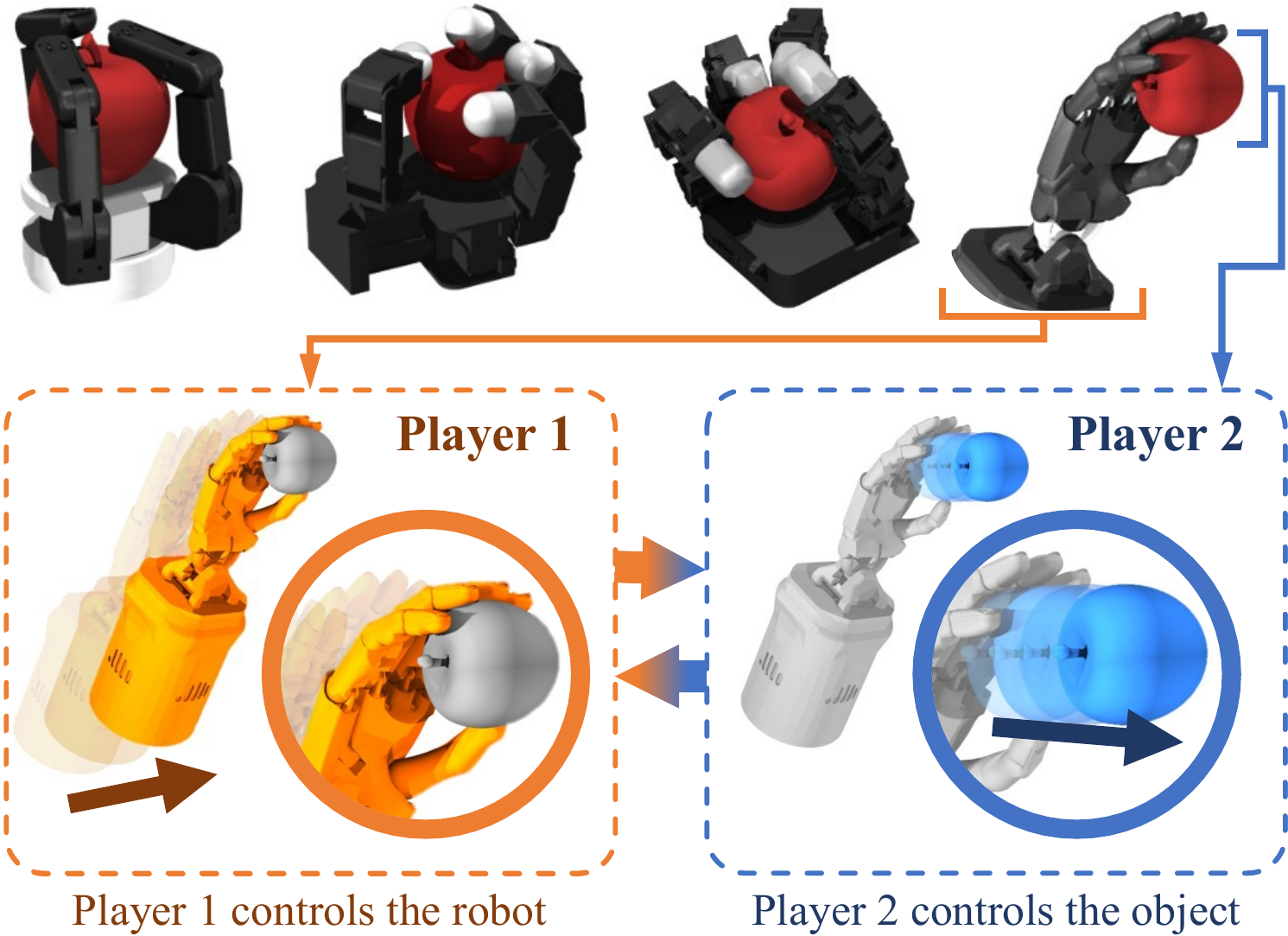}
    \caption{Our approach address the grasp synthesis problem for multi-fingered robotic hands by formulating it as a two-player game: Player~1 controls the hand to generate feasible grasp configurations, while Player~2 adversarially controls the object to attempt escape.}
    \label{fig:cover}
\end{figure}

To address this challenge, we propose a new grasp synthesis approach 
% \jiayuan{a problem formulation and a new approach?} 
that generates reliable grasps within seconds.
In contrast to previous methods, our approach explicitly captures and leverages adversarial object motions during grasp generation. 
Our key insight is to treat grasping as a two-player game (Fig.~\ref{fig:cover}) in which each player solves its own constrained optimization problem. 
Player~1 controls the robot to generate grasp configurations, while Player~2 adversarially controls the object to attempt to escape.  
% New optimization problem formulations for the two players is also proposed by introducing a kinematic firm-grasp condition that couples the two problems. 
% \jiayuan{To efficiently solve this minimax problem? Or at the implementational level?} 
% We propose a novel optimization problem formulation that couples the two players' problems by introducing a kinematic firm-grasp condition. 
At the implementation level, we formulate a novel optimization problem that couples the objectives of the two players through a kinematic firm-grasp condition.
Player~1 is constrained to satisfy this condition, while Player~2 is constrained to break it. 
% By taking the possible adversarial object motion into consideration, our approach constrains the object more effectively, thereby enhancing the overall robustness of the grasp. 
By taking the possible adversarial object motion into consideration, our approach effectively constrains the object escape, thereby enhancing the overall robustness of the grasp. 
% Our technical contributions are twofold: (i) a novel two-player game formulation of the grasp synthesis problem, and (ii) an algorithm that solves this game via an iterative best response strategy, with each player's optimization problem addressed using the Augmented Lagrangian method.

We evaluate our approach extensively in both simulation and real-world experiments. The simulation results on diverse robotic platforms and object sets demonstrate that our method achieves a grasp success rate (i.e., generating a grasp that prevents objects from escaping under external forces) of up to $19.61\%$ higher than the state-of-the-art baseline under external disturbance forces applied to the object.
% The effectiveness of the two-player game is further suggested by a success rate improvement of $27.40\%$ on average over the method that optimizes only the robot configuration. 
The effectiveness of the two-player game is further suggested by a success rate improvement of $27.40\%$ on average over the method without the game formulation, where when we only optimize for the grasp without considering external disturbance.
% \jiayuan{it's unclear what do you mean by "only the optimization for player-1 is activated." maybe say, when we only optimize for the grasp under a uniform external disturbance distribution in contrast to an adversarial one.}
% Moreover, our approach consistently generates stable grasps even when the object undergoes dramatic shape changes due to its robustness, which is hard to achieve for baseline approaches. 
% Further, we conduct an extreme case study to test the robustness and adaptability of our approach in challenging scenarios \jiayuan{I am not sure if the proposed challenging scenario is really about "stress testing" the approach... Is this more like a continuous property of the algorithm's generation?}, in which the proposed approach generates grasps online for an object that actively changes its shape over time.
Further, we conduct an extreme case study to test the robustness and adaptability of our approach, in which the proposed approach generates grasps online for an object that actively changes its shape over time.
% The results show that the robot hand is able to sustain stable grasps throughout the entire transformation process without failure.
The results show that the robot hand is able to sustain stable grasps throughout the entire shape-changing process without failure.
Our approach generates grasps efficiently, requiring 0.26\,s on Barrett (8 DoF), 0.78\,s on Allegro (16 DoF), and 1.04\,s on ShadowHand (24 DoF). 
% \jiayuan{and achieves a real-world success rate of XX on XX, YY on YY.}
% Real-world experiments on LeapHand and ShadowHand further shows the feasibility and effectiveness of our formulation.
Our real-world evaluations show success rates of $85.0\%$ on the ShadowHand and $87.5\%$ on the LeapHand, which further shows the feasibility and effectiveness of our formulation. 

In summary, our technical contributions are twofold: 
\begin{itemize}
    \item A novel two-player game formulation of the grasp synthesis problem;
    \item An efficient algorithm that solves this game via an iterative best response, with each player's optimization problem addressed using the Augmented Lagrangian method. 
    % \jiayuan{do you want to talk about your point-based representation for the grippers and the target objects?}
\end{itemize}

\section{Related Works}

% \subsection{Grasp Synthesis for Dexterous Robot Hands}
\subsection{Adversarial Approaches in Dexterous Grasp Synthesis}

Adversarial approaches have recently been applied to dexterous grasping, primarily as a means to assess grasp robustness and filter out insufficiently robust grasps. 
Recent studies on adversarial attacks on robot grasps similarly probe robustness, although they do not directly address grasp-pose optimization itself \cite{wang2025advgrasp}. 
Examples include adversarial objects deliberately designed to induce failures and stress-test grasping pipelines \cite{wang2019adversarial}, as well as feature-level adversarial training approaches that improve cross-domain grasp detection without modifying the pose solver \cite{jing2021domain}.
% Another line of work introduces execution-time adversarial perturbations to diagnose grasp stability and trigger corrective control. These perturbations include external forces or controlled deformations. For example, \cite{dong2019maintaining} employs vision-based tactile sensing to estimate contact deformations and slips.

\subsection{Conditions for Firm Grasp}
The conditions that determine whether a grasp constrains the target object form the foundation of grasp-synthesis analysis. 
Some firm grasp conditions are based on kinematic constraints that preclude any geometrically feasible motion of the object.
Form closure \cite{ferrari1992planning} is one of the best-known kinematic firm grasp conditions. It defines a grasp in which no object motion is feasible without violating contact constraints \cite{bicchi1995closure, markenscoff1990geometry}. 
Another kinematic firm grasp condition known as \textit{caging} confines the object to the space formed with bounded connected components \cite{dong2024cagecoopt, rimon1999caging}.
The condition proposed by our method shares a similar philosophy to form closure. The key difference is that, instead of analyzing eigenvectors at the contact points, our condition relies on a differential collision indicator function. 
This brings us two advantages:
First, the differential function can be directly incorporated into local optimization algorithms based on gradients.
Second, it does not rely on explicitly identifying contact points, which makes the approach more robust to contact uncertainty.
An alternative category is the force closure \cite{prattichizzo2008handbook}, which verifies whether contact wrenches counteract any external wrench applied to the object.

\begin{figure*}[t]
     \centering
    \includegraphics[width=0.95\linewidth]{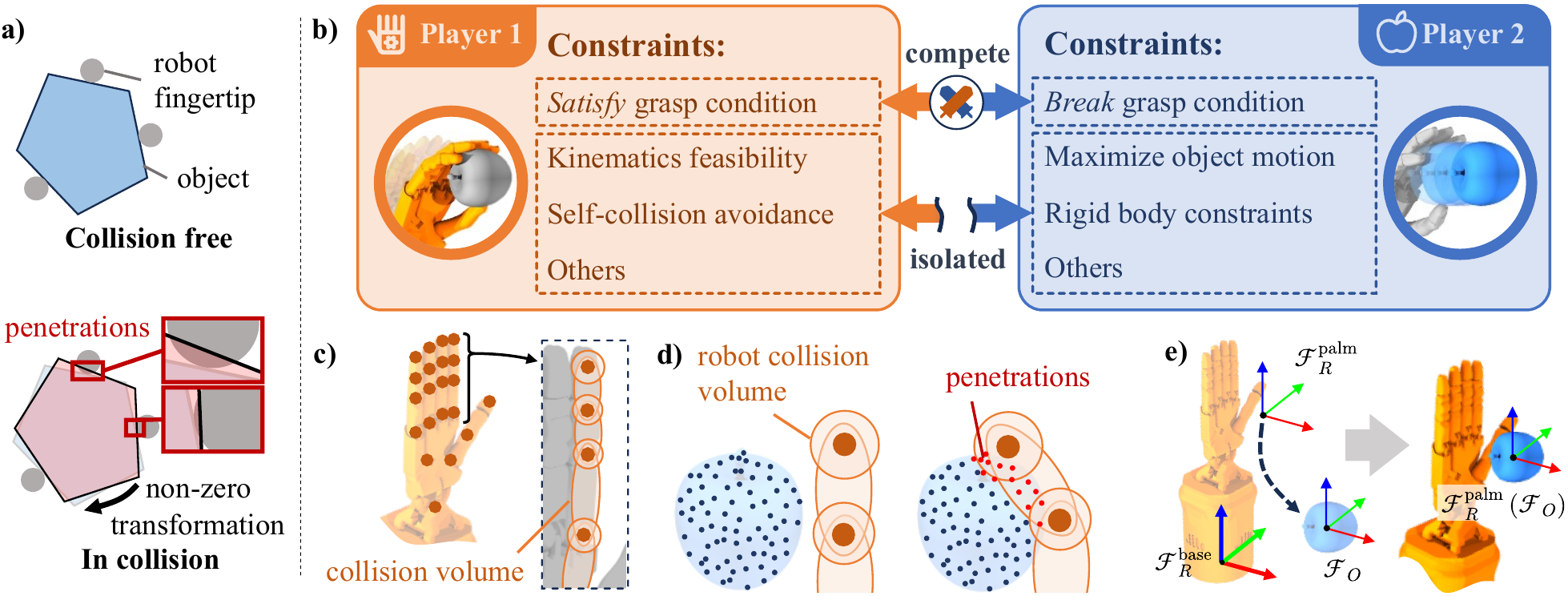}
    \caption{Overview of the proposed grasp synthesis approach. a) Our method relies on a firm grasp condition that once satisfied, any non-zero object transformation will result in object-robot penetration. b) The problem is formulated as a two-player game. Player~1 seeks to satisfy the firm grasp condition, while Player~2 attempts to break it. The two players compete specifically on this condition, whereas all other constraints are isolated within their respective optimization problems. c) We model the hand by attaching spatial points to base and joint frames: joints are represented as spheres at joint centers, and links as ellipsoids with foci at their endpoints. d) The object is modeled as a set of points directly sampled from its point cloud. Penetration occurs if any of these points fall within the robot’s collision volume. e) For algorithm initialization, we align the palm frame $\mathcal{F}_R^\text{palm}$ and the object frame $\mathcal{F}_O$.}
    \label{fig:pipeline}
\end{figure*}

\section{Grasp Synthesis as a Two-Player Game}
We focus on a floating-base multi-finger robot hand with $M$ joints. The robot state is denoted as $\boldsymbol{x}_R \in \mathbb{R}^{M} \times SE(3)$, where the robot base pose is represented using a three-dimensional translation and rotation. 
% parameterized by Euler angles.
We first introduce the firm grasp condition in Sec.~\ref{subsec:firm-grasp-condition}. 
Based on the firm grasp condition, we formulate the grasping synthesis problem as a two-player game in Sec.~\ref{subsec:game-formulation}. 
Finally, the objective and constraints that appeared in the optimization problems are detailed in Sec.~\ref{sec:objective-and-constraints}.

\subsection{Firm Grasp Condition}\label{subsec:firm-grasp-condition}
We consider the firm grasp condition that the robot hand kinematically constrains a rigid object. 
% \jiayuan{Mention "form closure?"} 
In this setting, the object is not allowed to undergo any further infinitesimal motion without colliding with the robot hand. An illustrative example is shown in Fig.~\ref{fig:pipeline}a. 
Assuming that the object local frame is initially aligned with the world frame, the condition can be expressed as:
\vspace{-2mm}
\begin{subequations} \label{eq:phi_conditions}
\begin{align}
&\phi(\boldsymbol{x}_R, \boldsymbol{I}) \geq 0, \label{eq:phi_1a}  \\
&\phi(\boldsymbol{x}_R, \exp{(\hat{\boldsymbol{\delta}}))} < 0, \; \forall \, \boldsymbol{\delta} \; \text{with} \; \lvert \boldsymbol{\delta} \rvert < \boldsymbol{\epsilon}. \label{eq:phi_1b}
\end{align}
\end{subequations}
% <- don't start a new paragraph
Here, $\phi(\cdot, \cdot)$ denotes a collision indicator function that takes robot state and object pose as input and describes whether the object and the robot hand penetrate. $\phi \geq 0$ indicates a collision-free and $\phi<0$ indicates penetration. 
% \jiayuan{wait, why is a collision detection function differentiable? Maybe you should give a forward pointer? Actually, the choice of using a differentiable indicator function here is for the algorithm, as in whether it's differentiable doesn't matter at the problem formulation level.}
The variable $\boldsymbol{\delta} \in \mathfrak{se}(3)$ represents a twist and $\exp(\hat{\boldsymbol{\delta}}) \in \mathrm{SE}(3)$ is the rigid-body transformation applied to the object.
Eq.~\eqref{eq:phi_1a} ensures that the object is not in collision with the robot hand in its initial (untransformed) state. Eq.~\eqref{eq:phi_1b} states that any small non-zero transformation of the object leads to a collision with the grasper, thereby implying that the object motion is fully constrained.

In practice, we bound the absolute element value of the variable $\boldsymbol{\delta}$ by a small positive $\boldsymbol{\epsilon} > 0$ due to two reasons: 
First, it avoids degenerate cases caused by object symmetries. 
For example, when grasping highly symmetric objects such as cylinders or cubes, a large motion (e.g., a $180^\circ$ rotation around the symmetry axis) may not change the contact geometry and thus would not trigger a collision. 
Without the small twist magnitude boundary, such symmetry-induced motions could be incorrectly interpreted as valid escape behaviors, causing the firm grasp condition to fail.
Second, it narrows the search space, which facilitates faster convergence of the subsequent local optimization algorithm.

\subsection{Two-Player Game Formulation}\label{subsec:game-formulation}
% \jiayuan{Maybe one or two sentences saying why do you want to solve this problem via a two-player game formulation? Why can't you just direclty solve the previous formulation.}
Directly seeks a robot state  $\boldsymbol{x}_R$ that satisfies firm grasp condition in Eqs.~\eqref{eq:phi_1a},~\eqref{eq:phi_1b} is challenging due to the presence of infinite constraints as the condition must hold for all $\boldsymbol{\delta}$.
Therefore, we model the grasp-synthesis problem as a two-player game (Fig.~\ref{fig:pipeline}b):
\begin{itemize}
    \item Player~1 seeks a robot state $\boldsymbol{x}_R^*$ that satisfies both Eq.~\eqref{eq:phi_1a} and Eq.~\eqref{eq:phi_1b} for any admissible object transformation represented by $\boldsymbol{\delta}$. 
    \item Player~2 seeks a transformation represented by $\boldsymbol{\delta}^*$ that displaces the object from its initial state without colliding with the robot hand. Equivalently, Player~2 aims to \emph{violate} the firm grasp constraint in Eq.~\eqref{eq:phi_1b}.
\end{itemize}

The behavior of the players is modeled by assuming that each player chooses inputs to solve its own constrained optimization problem:

{\small
\vspace{-3mm}
\begin{tabular}[h]{ll}
\begin{minipage}[t]{0.48\linewidth}
\begin{equation} 
\label{eq:player1}
\begin{aligned}
&\min_{\boldsymbol{x}_R} \quad J(\boldsymbol{x}_R) \\
&\text{ s.t. } q_1\left( \boldsymbol{x}_R \right) = 0, \\
&\qquad \boldsymbol{x}_R^- \leq \boldsymbol{x}_R \leq \boldsymbol{x}_R^+, \\
&\qquad \phi\left( \boldsymbol{x}_R, \exp{(\hat{\boldsymbol{\delta}})} \right) \leq 0, \\
&\qquad \phi\left( \boldsymbol{x}_R, \boldsymbol{I} \right) \geq 0, \\
&\qquad \phi^\text{self}\left( \boldsymbol{x}_R \right) \geq 0
\end{aligned}
\tag{P1}
\end{equation}
\end{minipage}
&
\begin{minipage}[t]{0.39\linewidth}
\begin{equation} 
\label{eq:player2}
\begin{aligned}
&\max_{\boldsymbol{\delta}} \quad \lVert\boldsymbol{\delta}\rVert_2^2 \\
&\text{ s.t. } q_2\left( \boldsymbol{\delta} \right) = 0, \\
&\qquad -\boldsymbol{\epsilon} < \boldsymbol{\delta} < \boldsymbol{\epsilon}, \\
&\qquad \phi\left( \boldsymbol{x}_R, \exp{(\hat{\boldsymbol{\delta}})} \right) \geq 0
\end{aligned}
\tag{P2}
\end{equation}
\end{minipage}
% \vspace{3mm}
\end{tabular}
}

Player~1 solves the optimization problem defined in \eqref{eq:player1}, which minimizes the objective function $J$ subject to a set of constraints. 
$q_1 = 0$ denotes the kinematics feasibility constraints.
We also consider the limits of the robot state as box constraints $\boldsymbol{x}_R^- \leq \boldsymbol{x}_R \leq \boldsymbol{x}_R^+$, where $\boldsymbol{x}_R^-$ and $\boldsymbol{x}_R^+$ are the lower and upper limits of the robot states. 
We use $ \phi^\text{self}\left( \boldsymbol{x}_R \right) \geq 0$ to denote self-collision-avoidance constraint of the robot.
In addition, Player~1 must satisfy the firm grasp constraints specified in Eq.~\eqref{eq:phi_1a} and Eq.~\eqref{eq:phi_1b}. 
Player~2 solves the optimization problem given in \eqref{eq:player2}, which maximizes the squared norm of the object transformation. 
The constraints include object point cloud constraints $q_2 = 0$, object transformation boundary $-\boldsymbol{\epsilon} < \boldsymbol{\delta} < \boldsymbol{\epsilon}$, and collision-free constraints $
\phi\left( \boldsymbol{x}_R, \boldsymbol{\delta} \right) \geq 0$. 
% \jiayuan{Say one sentence that it can be proved that the Nash equilibrium of this game corresponds to an optimal solution to the original formulation.}
% The formulations of the objectives and constraints are introduced in Sec.~\ref{sec:objective-and-constraints}.

\subsection{Objectives and Constraints} \label{sec:objective-and-constraints}
To compute the objective and constraints in \eqref{eq:player1} and \eqref{eq:player2}, 
we adopt the approach of Chen et al.~\cite{chen2024propagative} and represent the object and the robot as two point sets consisting of $N$ and $K$ points, respectively.
As shown in Fig.~\ref{fig:pipeline}c, the robot point set consists of points attached to the robot’s base and joint frames to capture the transformation of all rigid bodies.
As shown in Fig.~\ref{fig:pipeline}d, the object point set is sampled directly from the object point cloud. 
The position of the object and the robot points are denoted as $\boldsymbol{P}_O = \left[\boldsymbol{p}_{O_1}, \boldsymbol{p}_{O_2}, \ldots, \boldsymbol{p}_{O_N}\right]$ and $\boldsymbol{P}_R = \left[\boldsymbol{p}_{R_1}, \boldsymbol{p}_{R_2}, \ldots, \boldsymbol{p}_{R_K}\right]$. 
The objective functions and constraints are then formulated in terms of the pairwise distances between these points. 
The formulation of the robot kinematics constraints, the robot states constraints, and the constraints related to robot-object collisions have already been introduced in \cite{chen2024propagative}. 
The self-collision avoidance constraints for the robot and the object-related constraints are newly introduced in this paper.

\subsubsection{Robot kinematics constraints}
We use $q_1\left( \boldsymbol{x}_R \right) = 0$ to describe the kinematics relationship that maps the robot state $\boldsymbol{x}_R$ to its respective set of points $\boldsymbol{P}_R$:

{\small
\vspace{-1mm}
\begin{equation}
    \boldsymbol{P}_R = FK(\boldsymbol{x}_R), \label{eq:fk}
\vspace{-1mm}
\end{equation}
}
where $FK$ represents forward kinematics formulated by modifying the proximal \ac{DH} convention \cite{craig2006introduction} in terms of the distances between the spatial points attached to the robot hand.
Instead of using joint angles, we use these distances between the point sets to represent robot states $\boldsymbol{x}_R$.
This formulation eliminates the trigonometric terms in the mapping from robot states to point sets and reduces the risk of convergence to local minima when solving \ac{IK} with iterative local optimization methods~\cite{chen2024propagative}.

\subsubsection{Object point cloud constraints}
We use $q_2\left( \boldsymbol{\delta} \right) = 0$ to constrain the position of the object points:
{\small
\vspace{-1mm}
\begin{equation}
    q_2\left( \boldsymbol{\delta} \right) =
    \boldsymbol{P}_O 
    - \left( \exp\!(\hat{\boldsymbol{\delta}}) 
    \begin{bmatrix} 
        \boldsymbol{P}_O^{\text{ini}}, \; 
        1 
    \end{bmatrix}^\top \right)_{1:3} = 0,
    \label{eq:obj_tsfmt}
\vspace{-1mm}
\end{equation}
}
% where $T(\boldsymbol{\delta})$ denotes the rigid-body spatial transformation induced by $\boldsymbol{\delta}$, and $\boldsymbol{P}_O^{\text{ini}}$ represents the initial positions of the object points corresponding to $\boldsymbol{\delta}=0$.
where $\boldsymbol{P}_O^{\text{ini}}$ represents the initial positions of the object points corresponding to $\boldsymbol{\delta}=0$. 
% \jiayuan{This essentially "fixes" the object at the origin.}

\subsubsection{Box constraints on robot states}\label{sec:box-constraint-robot}
We use inequality constraints $C(\boldsymbol{x}_R) \leq 0$ to represent the limits of the robot state.
In this work, we focus on box constraints
$\boldsymbol{x}_R^- \leq \boldsymbol{x}_R \leq \boldsymbol{x}_R^+$ that arise from joint angle limits and robot base transformation limits.
To ensure robot state limits by construction, we convert box constraints to equality constraints employing the sigmoid function $\sigma$ 
% \cite{han1995influence} 
and introduce an additional decision variable $\boldsymbol{\omega}_R \in \mathbb{R}^{6+M} \times SE(3)$:  
\vspace{-1mm}
\begin{equation}
    \boldsymbol{x}_R = s_R(\boldsymbol{\omega}_R) = \left(\boldsymbol{x}_R^+ - \boldsymbol{x}_R^-\right) \, \sigma(\boldsymbol{\omega}_R) + \boldsymbol{x}_R^-,
    \label{eq:squash1}
\vspace{-1mm}
\end{equation}
where $\sigma(\boldsymbol{\omega}_R) = 1 / \left( 1 + e^{-\boldsymbol{\omega}_R}\right)$.
The value of the sigmoid function is bounded within $(0, 1)$, which ensures that the robot states are confined within $(\boldsymbol{x}_R^-, \boldsymbol{x}_R^+)$, providing a close approximation to the closed interval $\left[ \boldsymbol{x}_R^-, \boldsymbol{x}_R^+ \right]$. 

\subsubsection{Box constraints on object transformations}
Similar to robot state constraints, we apply Sigmoid function and additional decision variable $\boldsymbol{\omega}_O \in \mathbb{R}^6$ to convert the object transformation limits to equality constraints:
\vspace{-1mm}
\begin{equation}
    \boldsymbol{\delta} = s_O(\boldsymbol{\omega}_O) = 2\boldsymbol{\epsilon} \, \sigma(\boldsymbol{\omega}_O) - \boldsymbol{\epsilon}.
    \label{eq:squash2}
\vspace{-1mm}
\end{equation}

\subsubsection{Differentiable constraints related to robot-object collisions}
Collisions are detected by computing the distances between points in the robot point set and the object point set, and comparing the distances against a prescribed threshold. For the $k$-th point attached to the robot and the $n$-th point attached to the object, we use function $\psi$ to detect collisions
\vspace{-1mm}
\begin{equation}
    \psi^{k,n}(\boldsymbol{P}_R, \boldsymbol{P}_O) =  ||\boldsymbol{p}_{R_k} - \boldsymbol{p}_{O_n}||_2 - r_k,
    \label{eq:joint_coll_avoid}
\vspace{-1mm}
\end{equation}
where $r_k$ is the distance threshold for the point $\boldsymbol{p}_{Rk}$.
\vspace{1mm}

We also consider the collision detection between the robot link and the object. If there is a rigid link connecting the robot point $\boldsymbol{p}_{Rk1}$ and $\boldsymbol{p}_{Rk2}$, 
{\small
\begin{equation}
    \psi^{k_1, k_2 ,n}(\boldsymbol{P}_R, \boldsymbol{P}_O) = ||\boldsymbol{p}_{Rk1} - \boldsymbol{p}_{On}||_2 + ||\boldsymbol{p}_{Rk2} - \boldsymbol{p}_{On}||_2 - r_{k1, k2} \; ,
    \label{eq:link_coll_avoid}
\end{equation}}
where $r_{k1, k2}$ is the distance threshold for the link that connects point $\boldsymbol{p}_{Rk1}$ and $\boldsymbol{p}_{Rk2}$. We summarize all $\psi^{k,n}$ and $\psi^{k_1, k_2 ,n}$ as $\psi(\boldsymbol{P}_R, \boldsymbol{P}_O)$:
\vspace{-1mm}
\begin{equation}
    \phi(\boldsymbol{x}_R, \boldsymbol{\delta}) = \psi(\boldsymbol{P}_R, \boldsymbol{P}_O). \label{eq:collision-indicator}
\vspace{-1mm}
\end{equation}

\subsubsection{Differentiable robot self-collision avoidance constraints}
% \jiayuan{Say it's differentiable!}
We formulate the self-collision avoidance constraints $\phi^\text{self}\left( \boldsymbol{x}_R \right) \geq 0$  by replacing the object point in Eq.~\eqref{eq:joint_coll_avoid} and Eq.~\eqref{eq:link_coll_avoid} with robot points:
\vspace{-1mm}
\begin{equation}
    \psi^{k,k'} (\boldsymbol{P}_R) =  ||\boldsymbol{p}_{Rk} - \boldsymbol{p}_{k'}||_2 - r_k, \; k' \neq k,
    \label{eq:self_joint_coll_avoid}
\vspace{-1mm}
\end{equation}
{\small
\vspace{-1mm}
\begin{equation}
\begin{aligned}
\psi^{k_1, k_2 ,k'} (\boldsymbol{P}_R) =
& ||\boldsymbol{p}_{Rk1} - \boldsymbol{p}_{Rk'}||_2 + ||\boldsymbol{p}_{Rk2} - \boldsymbol{p}_{Rk'}||_2 \\
& \; - r_{k1, k2}, \; k' \neq k_1 \text{ and } k' \neq k_2,
\end{aligned}
\label{eq:self_link_coll_avoid}
\vspace{-1mm}
\end{equation}
}
We summarize Eq.~\eqref{eq:self_joint_coll_avoid} and Eq.~\eqref{eq:self_link_coll_avoid} as $\psi^\text{self}(\boldsymbol{P}_R)$:
\vspace{-1mm}
\begin{equation}
    \phi^\text{self}\left( \boldsymbol{x}_R \right) = \psi^\text{self}(\boldsymbol{P}_R) \geq 0 \label{eq:self-collision-indicator}
\vspace{-1mm}
\end{equation}

\subsubsection{Grasping objective function}
We design a grasping objective function $J(\boldsymbol{x}_R)$ that encourages a designated subset of robot points, typically those attached to the robot's fingertips, to contact the object surface. 
Let $\boldsymbol{P}_R' = [\boldsymbol{p}_{R_1}', \boldsymbol{p}_{R_2}', \ldots, \boldsymbol{p}_{R_K'}']$ denote the positions of this subset, which are computed from the robot state $\boldsymbol{x}_R$ using Eq.~\eqref{eq:fk}. 

For the $k$-th point $\boldsymbol{p}_{R_k}'$, we consider it to be in contact with the object if the minimum distance between it and all object points is smaller than a threshold $r_k'$. 
We define this condition using

{\small
\vspace{-1mm}
\begin{equation}
    J_k(\boldsymbol{P}_R, \boldsymbol{P}_O) = \max \! \left(\min_{1 \leq n \leq N} \, \lVert \boldsymbol{p}_{R_k}' - \boldsymbol{p}_{O_n} \rVert_2 - r_k', \;0 \right)
\vspace{-1mm}
\end{equation}
}%
The grasping objective function is then defined as
{\small
\vspace{-1mm}
\begin{equation}
    J(\boldsymbol{x}_R) 
    = \frac{1}{2}\sum_{k=1}^{K'} {J_k(\boldsymbol{P}_R, \boldsymbol{P}_O)}^2,
     \label{eq:grasp-objective}
\vspace{-1mm}
\end{equation}
}
which is formulated to drive each designated robot point $\boldsymbol{p}_{Rk}'$ toward its closest object point, thereby encouraging fingertip contact with the object surface.

\section{Algorithm}
% \jiayuan{Give an overview. At a high-level, our algorithm illustrated in XX, is based on iterative best-response. At each iteration, two players take turns to optimize for its own decision variable given the decision variable of the other player fixed. In the following, we first introduce X in X, Y in Y, Z in Z...}
Our algorithm solves the game using an iterative best-response strategy.
At each iteration, the two players alternately optimize their own decision variable while keeping the other fixed.
We first describe how to solve the inner optimization problems for each player in Sec.~\ref{sec:inner-problem}, and then present the overall game solution in Sec.~\ref{sec:ibr}.
As illustrated in Fig.~\ref{fig:pipeline}e, the relative pose between the robot hand and the object is initialized by defining a palm frame $\mathcal{F}_R^\text{palm}$ with a manually specified transformation with respect to the robot base frame $\mathcal{F}_R^\text{base}$. For initialization, $\mathcal{F}_R^\text{palm}$ is aligned with the object frame $\mathcal{F}_O$. The origin of $\mathcal{F}_O$ is placed at the geometric center of the object, while its axes are aligned with the principal directions obtained via \ac{PCA} on the object’s point cloud.

\subsection{Solving Inner Optimization Problems} \label{sec:inner-problem}
We first reformulate the optimization problem \eqref{eq:player1} and \eqref{eq:player2}. We represent the robot–object collision indicator function $\phi$, robot self-collision indicator function $\phi^\text{self}$, and grasping objective $J$ using the decision variable $\boldsymbol{\omega}_R$ and $\boldsymbol{\omega}_O$. This is achieved by incorporating Eqs.~\eqref{eq:fk}, \eqref{eq:obj_tsfmt}, \eqref{eq:squash1}, and ~\eqref{eq:squash2} into Eqs.~\eqref{eq:collision-indicator}, \eqref{eq:self-collision-indicator}, and \eqref{eq:grasp-objective}:  

{\small
\begin{equation}
    \phi(\boldsymbol{x}_R, \boldsymbol{\delta}) 
    = \phi'(\boldsymbol{\omega}_R, \boldsymbol{\omega}_O) 
    = \psi\!\left(FK(s(\boldsymbol{\omega}_R)), \; T(s(\boldsymbol{\omega}_O)) \, \boldsymbol{P}_O^{\text{ini}}\right),
\end{equation}
\begin{equation}
    \phi^\text{self}\left( \boldsymbol{x}_R \right) 
    = {\phi^\text{self}}'\left( \boldsymbol{\omega}_R \right) 
    = \psi^\text{self}(FK(s(\boldsymbol{\omega}_R))),
\end{equation}
\begin{equation}
    J(\boldsymbol{x}_R) 
    = J'(\boldsymbol{\omega}_R) 
    = \sum_{k=1}^{K'} \frac{J_k\left(FK(s(\boldsymbol{\omega}_R)), \; T(s(\boldsymbol{\omega}_O)) \, \boldsymbol{P}_O^{\text{ini}}\right)^2}{2}.
% \vspace{-1mm}
\end{equation}
}

To enforce the robot state and object transformation limits by construction, the optimization for \eqref{eq:player1} is carried out over $\boldsymbol{\omega}_R$ rather than $\boldsymbol{x}_R$, and the optimization for \eqref{eq:player2} is carried out over $\boldsymbol{\omega}_O$ rather than $\boldsymbol{\delta}$.

The reformulated optimization problems are thus given by  

{\small
\vspace{-2mm}
\begin{tabular}[h]{ll}
\begin{minipage}[t]{0.45\linewidth}
\begin{equation} 
\label{eq:player1-converted}
\begin{aligned}
&\min_{\boldsymbol{\omega}_R} \quad J'(\boldsymbol{\omega}_R)  \\
&\text{ s.t. } C(\boldsymbol{\omega}_R, \boldsymbol{\omega}_O) \leq 0,
\end{aligned}
\tag{P1$^\dagger$}
\end{equation}
\end{minipage}
&
\begin{minipage}[t]{0.45\linewidth}
\begin{equation} 
\label{eq:player2-converted}
\begin{aligned}
&\max_{\boldsymbol{\omega}_O} \quad \lVert s(\boldsymbol{\omega}_O) \rVert_2^2 \\
&\text{ s.t. } \phi'(\boldsymbol{\omega}_R, \boldsymbol{\omega}_O) \geq 0
\end{aligned}
\tag{P2$^\dagger$}
\end{equation}
\end{minipage}
\vspace{2mm}
\end{tabular}}
where {\small $C = 
\begin{bmatrix}
\phi'(\boldsymbol{\omega}_R, \boldsymbol{\omega}_O) , \;
-{\phi^\text{self}}'\left( \boldsymbol{\omega}_R \right) , \;
-\phi'(\boldsymbol{\omega}_R, \boldsymbol{0})
\end{bmatrix}^\top$}.

We solve \eqref{eq:player1-converted} and \eqref{eq:player2-converted} using Augmented Lagrangian method that offloads the constraints into objective functions:
\vspace{-1mm}
\begin{equation}
L_{\rho1} = J' + \boldsymbol{\mu}_1 \max(C, 0) + \frac{\rho_1}{2} \max(C, 0)^\top \max(C, 0)
\vspace{-1mm}
\end{equation}
\vspace{-1mm}
\begin{equation}
\begin{aligned}
L_{\rho2} &= -\lVert s(\boldsymbol{\omega}_O) \rVert_2^2 + \boldsymbol{\mu}_2 \max(-\phi', 0) \\
& \qquad \quad + \frac{\rho_2}{2} \max(-\phi', 0)^\top \max(-\phi', 0)
\end{aligned}
\vspace{-1mm}
\end{equation}
where $\boldsymbol{\mu}_1$ and $\boldsymbol{\mu}_2$ are the Lagrangian multipliers. $\rho_1$ and $\rho_2$ are adjust penalty parameter. 
\eqref{eq:player1-converted} and \eqref{eq:player2-converted} are then converted into

\vspace{-3mm}
\begin{tabular}[h]{ll}
\begin{minipage}[t]{0.45\linewidth}
\begin{equation}
    \boldsymbol{\omega}_R^* = \arg\min L_{\rho1}
    \label{eq:solve-player1-ibr}
\end{equation}
\end{minipage}
&
\begin{minipage}[t]{0.45\linewidth}
\begin{equation}
    \boldsymbol{\omega}_O^* = \arg\min L_{\rho2}
    \label{eq:solve-player2-ibr}
\end{equation}
\end{minipage}
\vspace{3mm}
\end{tabular}

For player 1, we iteratively minimize $L_{\rho1}$. 
$\boldsymbol{\mu}_1$ and $\rho_1$ are initialized as $\boldsymbol{0}$ and $1$. 
Within each iteration, $\boldsymbol{\mu}_1$ and $\rho_1$ are fixed, and the \ac{L-BFGS} solver \cite{dennis1977quasi} is used to minimize $L_{\rho1}$. 
After obtaining the minimizer $\boldsymbol{\omega}_R^*$, we update $\boldsymbol{\mu}_1$ and scale $\rho_1$ by the factor $\alpha=10$. 
The process continues until the violation of the terminal state constraint $C$ falls below a tolerance $C_{\rm tol}$, the relative improvement becomes smaller than $r=10^{-2}$, or the number of iterations reaches $W_1$. 
Finally, the robot state $\boldsymbol{x}_R^*$ is recovered from $\boldsymbol{\omega}_R^*$ using Eq.~\eqref{eq:squash1}.  

A similar procedure is applied to Player~2.  
We minimize $L_{\rho2}$ and deal with $\boldsymbol{\mu}_2$ and $\rho_2$ in the same way as $L_{\rho1}$, $\boldsymbol{\mu}_1$, and $\rho_1$.
The loop ends when $\max(-\phi,0)$ is reduced below the tolerance $\phi_{\rm tol}$, its relative improvement is smaller than $r=10^{-2}$, or the number of iterations reaches $W_2$. 
Once the loop terminates, the object transformation $\boldsymbol{\delta}^*$ is obtained from $\boldsymbol{\omega}_O^*$ using Eq.~\eqref{eq:squash2}. 

\begin{figure}
     \centering
    \includegraphics[width=0.95\linewidth]{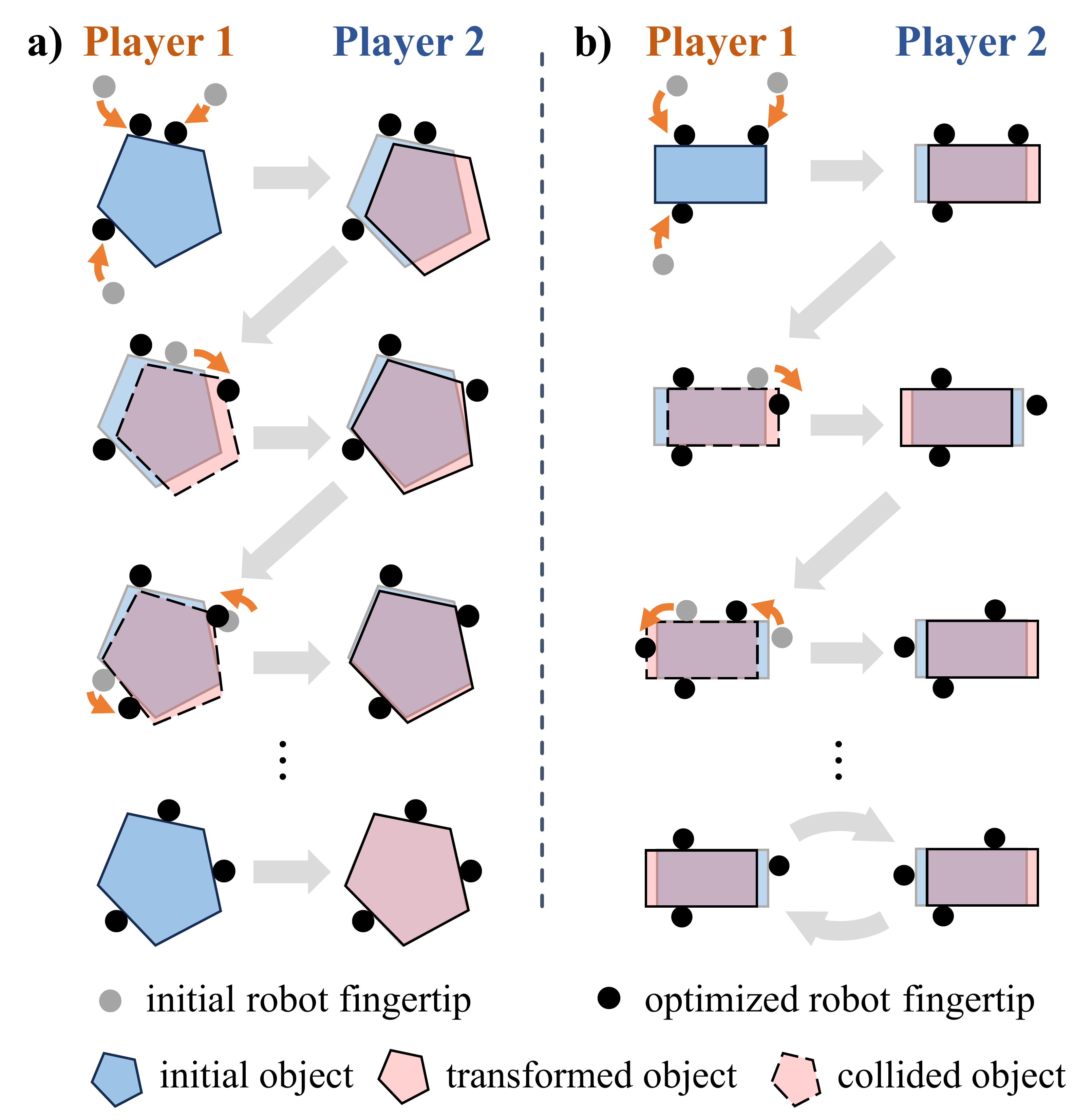}
    \caption{Visualization of convergence behaviors under the iterative best-response strategy. a) The algorithm converges to a fixed equilibrium point, where the robot hand successfully constrains the motion of the object. b) In other cases, the player interactions are trapped in a cyclic pattern.}
    \label{fig:converge-analysis}
\end{figure}

\subsection{Solving the Game via Iterative Best Response Strategy}\label{sec:ibr}

\begin{algorithm}[t]
{\small
\caption{Two-Player Iterative Best Response}
\label{alg:ibr}
\begin{algorithmic}[1]
\Require Player1, Player2, $\varepsilon$, $\varepsilon'$, $T$
\State $\boldsymbol{\delta}^0 \gets 0$
\For{$t = 1, \dots, T$}
    \State Player~1 updates $\boldsymbol{x}_R^{t}$ via Eq.~\eqref{eq:solve-player1-ibr} given $\boldsymbol{\delta}^{t-1}$.
    \State Player~2 updates $\boldsymbol{\delta}^{t}$ via Eq.~\ref{eq:solve-player2-ibr} given $\boldsymbol{x}_R^{t}$.
    \If{$\|\boldsymbol{\delta}^{t}\| < \varepsilon$ \textbf{or} $\|\boldsymbol{\delta}^{t}\| - \|\boldsymbol{\delta}^{t-1}\| < \varepsilon'$}
        \State $\boldsymbol{x}_R^{*} \gets \boldsymbol{x}_R^{t}$
        \State \textbf{break} %\Comment{Converged or trapped in a cycle}
    \EndIf
\EndFor
\State \Return $\boldsymbol{x}_R^{*}$
\end{algorithmic}
}
\end{algorithm}

\begin{figure*}[t]
    \centering
    \includegraphics[width=1\linewidth]{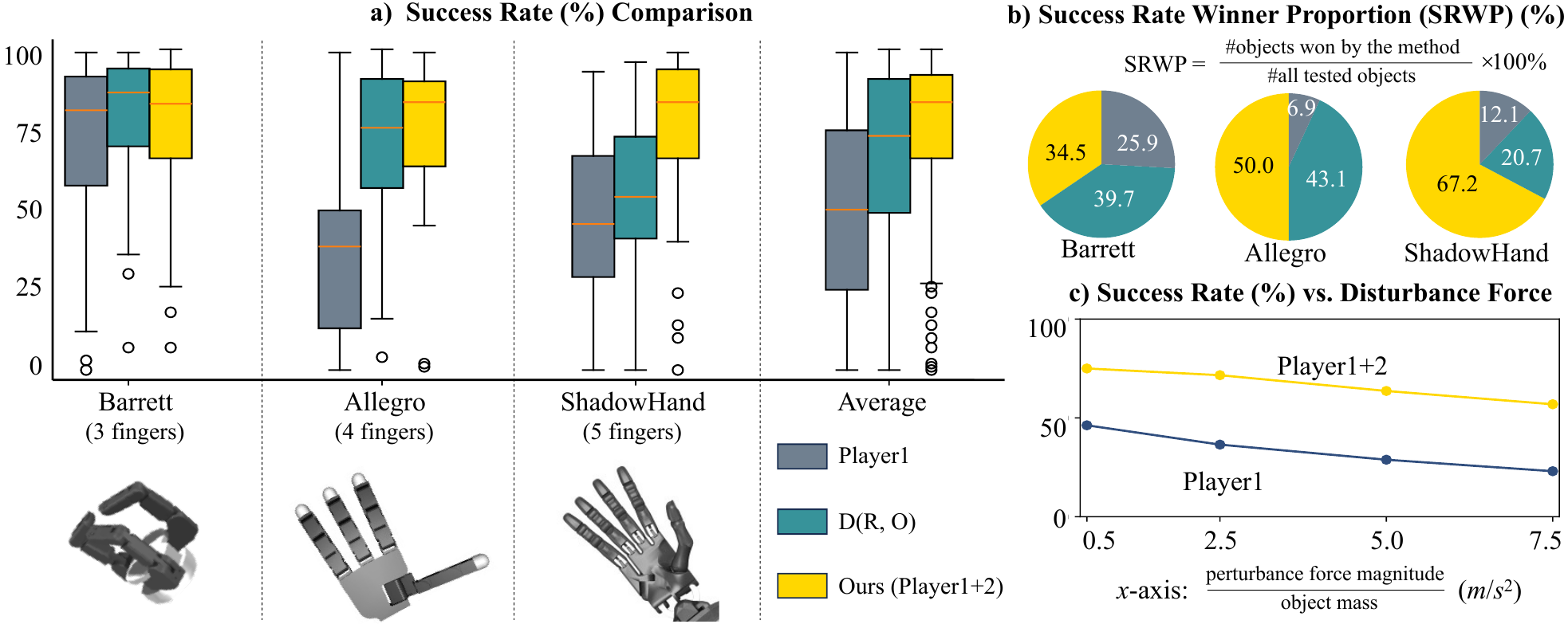}
    \caption{
    Comparison of grasp success rates across three robotic hands using 58 objects from the CMapDataset full set. (a) Distribution of success rates for three methods on Barrett, Allegro, ShadowHand, and their overall average. (b) Proportion of objects in which each method achieves the highest success rate. (c) How success rates vary under increasing disturbances.
    }
    \label{fig:comparison}
\end{figure*}

As shown in Algorithm~\ref{alg:ibr}, we solve the two-player game using the Iterative Best Response strategy. The two players alternately solve their respective optimization problems based on the current state of the other player. In each iteration, Player~1 updates first, followed by Player~2, formally,
% \vspace{-3mm}
% \begin{tabular}[h]{ll}
% \begin{minipage}[t]{0.43\linewidth}
% \begin{equation}
%     \boldsymbol{\omega}_R^t = \arg\min L_{\rho1} |_{\boldsymbol{\delta}=\boldsymbol{\delta}^{t-1}},
% \end{equation}
% \end{minipage}
% &
% \begin{minipage}[t]{0.43\linewidth}
% \begin{equation}
%     \boldsymbol{\omega}_O^t = \arg\min L_{\rho2}|_{\boldsymbol{x}_R=\boldsymbol{x}_R^{t}},
% \end{equation}
% \end{minipage}
% \vspace{3mm}
% \end{tabular}
\vspace{-3mm}
\begin{equation}
    \boldsymbol{\omega}_R^t = \arg\min L_{\rho1} |_{\boldsymbol{\delta}=\boldsymbol{\delta}^{t-1}},
\end{equation}
\begin{equation}
    \boldsymbol{\omega}_O^t = \arg\min L_{\rho2}|_{\boldsymbol{x}_R=\boldsymbol{x}_R^{t}},
\end{equation}
where $t=1, 2, \ldots, T$ is the iteration index and $T$ is the maximum number of iteration. 
% The details of solving Eq.~\eqref{eq:player1-ibr} and Eq.~\eqref{eq:player2-ibr} will be introduced in Sec.~\ref{sec:inner-problem}.
The iterative process continues until convergence or until the number of iterations reaches $T$. In practice, we observe two possible behaviors:

\textit{Convergence to a stable grasp}. 
A firm grasp is reached when the optimal object transformation solved by Player~2 approaches zero, i.e., $\boldsymbol{\delta}^* \rightarrow 0$. 
This means that the object is kinematically constrained by the robot hand and no collision-free transformation can be found by Player~2.
In this case, the two inequality constraints imposed on Player~1, $\phi(\boldsymbol{x}_R,\boldsymbol{\delta})\leq0$ and $\phi(\boldsymbol{x}_R,\boldsymbol{0})\geq 0$ merge into an equality constraint at the boundary, namely $\phi(\boldsymbol{x}_R,\boldsymbol{0})=0$. 
As a result, Player~1’s optimization reduces to minimizing its objective subject to this collapsed set of constraints, and no further improvement is possible for Player~1 in response to Player~2. 
A visualization example is shown in Fig.~\ref{fig:converge-analysis}a.
% The algorithm therefore reaches a fixed point. \footnote{From this reasoning, it follows directly that the strategy pair $(\boldsymbol{x}_R^*,\,\boldsymbol{\delta}^*=0)$ constitutes a Nash equilibrium.}. A visualization example is shown in \textcolor{red}{Fig~\ref{TODO}}.

\textit{Cyclic behavior}. In some cases, the two players may enter a cycle where successive updates alternate between conflicting strategies without settling. 
% Such situations arise \jiayuan{A common reason is that...} when kinematically constraining the object is infeasible, or when one or both players fail to solve their respective optimization problems. 
Common reasons are that kinematically constraining the object is sometimes infeasible, or that one or both players fail to solve their respective optimization problems.
As an example shown in Fig.~\ref{fig:converge-analysis}b, the rectangular object alternates between a left position and a right position, while the robot is unable to restrict the motion of the object.

We monitor the norm of the object transformation $\|\boldsymbol{\delta}\|$. If no significant progress is observed over multiple iterations, the algorithm is considered to have converged to a fixed point or entered a cycle and the algorithm terminates early.

\section{Experiments and Results}
We evaluate our approach through simulation and real-world experiments across a variety of objects, embodiments, and challenging scenarios. 
We start with, in Sec.~\ref{sec:compare}, benchmarking on the full CMapDataset with 58 diverse objects and validating performance across three dexterous robot hands platforms. 
To further test robustness, we design a stress test evaluation in which the robot must maintain a grasp on an object that continuously morphs across drastically different categories in Sec.~\ref{sec:extreme}. 
Finally, we validate our method on the real-world LeapHand and ShadowHand platforms with real household objects under tabletop settings in Sec.~\ref{sec:real}.
For all experiments, we set the maximum loop iterations as $T=10$, $W_1=100$, and $W_2=100$ and tolerance $C_\text{tol}=10^{-5}$, $\phi_\text{tol}=10^{-5}$, $\varepsilon=10^{-3}$, and $\varepsilon'=10^{-5}$. 

\subsection{Our Approach vs. Baselines}\label{sec:compare}
\subsubsection{Setup}
We evaluate grasp success based on whether the grasp constrains the object position under disturbance forces. We conduct experiments on the CMapDataset~\cite{li2022gendexgrasp}, which contains 58 test objects. The experiment setup primarily follows \cite{wei2024d}: We validate our approach on three widely used dexterous hands in simulation: Barrett, Allegro, and ShadowHand\footnote{Barrett: \texttt{www.barrett.com}. Allegro: \texttt{www.allegrohand.com}. ShadowHand: \texttt{www.shadowrobot.com}.}. The experiments are implemented in the Isaac Gym simulator~\cite{makoviychuk2021isaac}. A simple grasp controller proposed by Wei et al.~\cite{wei2024d} is used to execute the predicted grasps. External forces along six orthogonal directions are applied to the object, each lasting 1 second. The magnitude of the external forces equal to 0.5 times of object mass. A grasp is considered successful if the object's resultant displacement remains below 2\,cm after all six perturbations. Each method is evaluated with 100 grasp trials per object using randomly sampled point clouds. The success rate is the proportion of successful grasps among all trials.

\subsubsection{Baselines}
We compare our approach with the current state-of-the-art method D(R,O)~\cite{wei2024d}. 
D(R,O) is a learning-based grasp synthesis method that requires training data to acquire grasping strategies.
Following \cite{wei2024d}, D(R,O) is trained on the training split (48 objects) of the CMapDataset. The remaining 10 objects form the validation set that are unseen during training. 
In contrast, our approach is training-free, and all 58 objects are novel to our method. From this perspective, the experiment setup is slightly biased in favor of D(R,O).

We additionally include a baseline to highlight the contribution of the game-theoretic formulation in our approach. In this baseline approach, Player~2 is removed, leaving only Player~1 to optimize with grasping objectives, collision-avoidance constraints, and kinematic feasibility constraints. We name this baseline as \textit{Player1}:
\begin{equation} 
\label{eq:baseline-player1}
\begin{aligned}
&\min_{\boldsymbol{x}_R} \quad J(\boldsymbol{x}_R) \\
&\text{ s.t. } q_1\left( \boldsymbol{x}_R \right) = 0, \; \boldsymbol{x}_R^- \leq \boldsymbol{x}_R \leq \boldsymbol{x}_R^+, \\
&\qquad \phi\left( \boldsymbol{x}_R, \boldsymbol{0} \right) \geq 0, \; \phi^\text{self}\left( \boldsymbol{x}_R \right) \geq 0,
\end{aligned}
\tag{\small Baseline: Player1}
\end{equation}

\begin{figure}[t]
    \centering
    \includegraphics[width=1\linewidth]{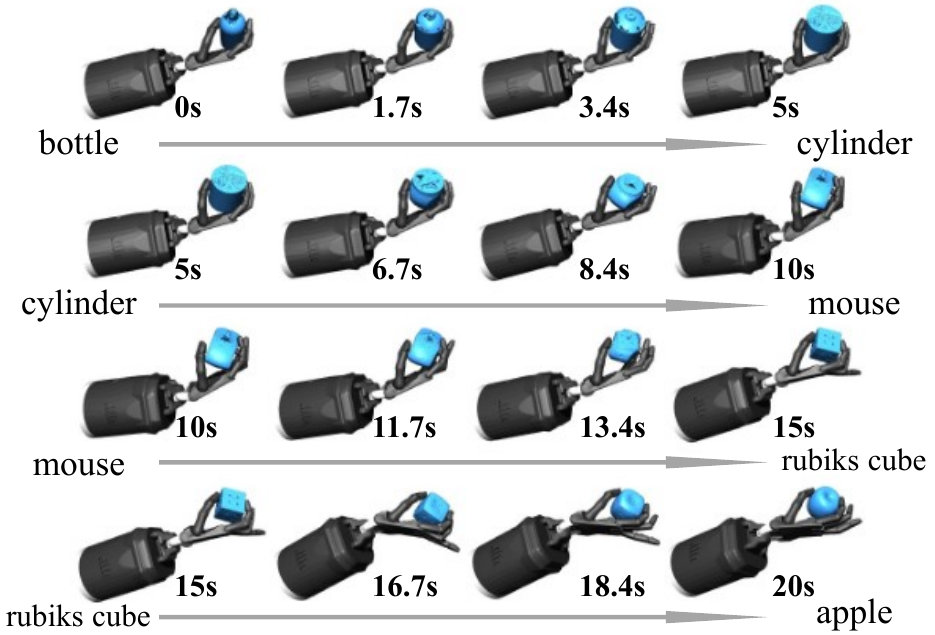}
    \caption{
    Key frames in extreme-case robustness evaluation. ShadowHand maintains a stable grasp while the target object continuously morphs across distinct categories.
    % : water bottle $\rightarrow$ cylinder $\rightarrow$ computer mouse $\rightarrow$ Rubik’s cube $\rightarrow$ apple. 
    % This challenging scenario shows the adaptability of our approach under severe cross-category shape deformations.
    }
    \label{fig:extreme}
\end{figure}

\subsubsection{Comparison Results}
As shown in Fig.~\ref{fig:comparison}a, the success rate of our approach significantly outperforms the baselines. 
Moreover, as the complexity of the robotic hand platform increases, the advantages of our method become more pronounced.
On the 3-finger Barrett hand, our approach achieves an average success rate of $76.59\%$, 
which is comparable to D(R,O) ($78.93\%$), with only a small difference of $2.34\%$. 
On the 4-finger Allegro hand, our success rate of $75.74\%$ surpasses D(R,O) 
% ($70.47\%$) 
by $5.27\%$ and Player1 
% ($35.00\%$) 
by $40.74\%$. 
On the 5-finger ShadowHand, our method achieves $75.02\%$, which is $19.61\%$ higher than 
D(R,O) 
% ($55.41\%$) 
and $34.28\%$ higher than Player1. 
% ($46.26\%$) 
On average, our approach achieves the grasp success rate of $75.78\%$, which is $7.51\%$ higher than D(R, O) and $25.56\%$ higher than Player~1.
These results demonstrate solution robustness and the adaptability of our approach for dexterous grasp synthesis.

We also compare the Success Rate Winner Proportion (SRWP) as shown in Fig.~\ref{fig:comparison}b.
SRWP is defined as follows: for each of the 58 objects, we evaluate all three methods and record the success rate of each. The method achieving the highest success rate for a given object is designated as the winner for that object. SRWP is then computed as the proportion of objects for which a method is the winner.
Our method achieves a comparable or higher SRWP compared to the baselines and the performance gap becomes more pronounced as hand dexterity increases.

To further evaluate the effectiveness of our two-player game formulation, we gradually increase the applied force on the object from $0.5$ to $7.5$ times its mass and record the corresponding success rates on the ShadowHand. As shown in Fig.~\ref{fig:comparison}c, our method consistently outperforms the single-player variant across all disturbance levels. It maintains significantly higher success rates even under strong perturbations, which confirms the robustness of our formulation.

\subsubsection{Runtime Analysis}
Our approach is implemented in C++ on a laptop computer with Core i7 CPU with 16GB RAM.
Our method takes an average runtime of 0.26s on Barrett, 0.78s on Allegro, and 1.04s on ShadowHand. The fast computation shows the feasibility of our approach in online dexterous manipulation tasks.

\subsection{Extreme-Case Robustness Evaluation: Grasping Cross-Category Deforming Object}\label{sec:extreme}

\begin{figure}[t]
    \centering
    \includegraphics[width=0.8\linewidth]{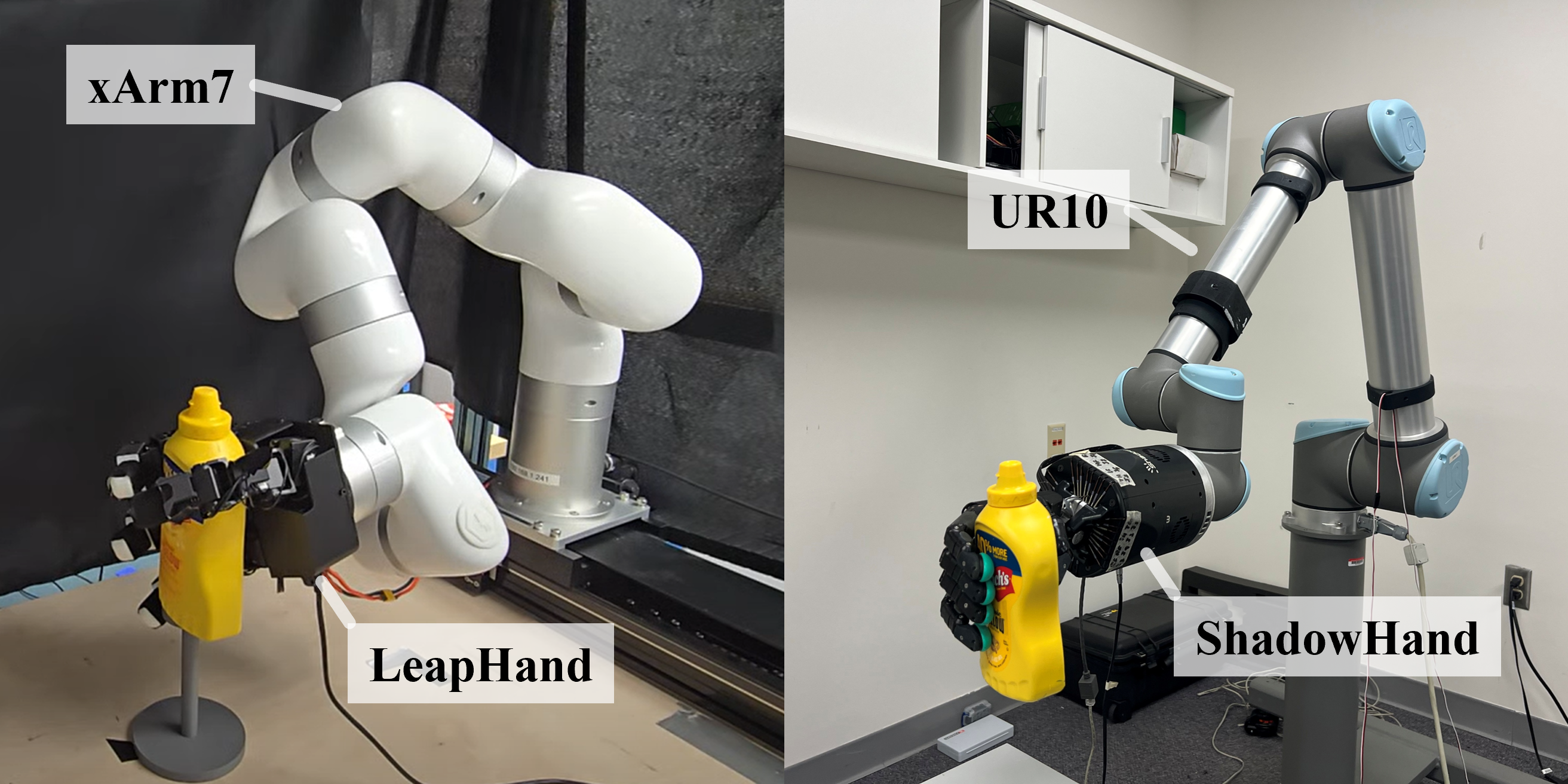}
    \caption{
    Real-world experiment setup.
    }
    \label{fig:real-world-exp}
\end{figure}

\begin{table}[h!]
    \caption{Real-World Experiment Results.}
    \centering
    \begin{tabular}{cccccc}
        \hline
        \textbf{Object} & tuna can & drill & tomato can & mustard & Avg. \\
        \hline
        \textbf{LeapHand} & 8/10 & 9/10 & 10/10 & 8/10 & 87.5$\%$ \\
        \textbf{ShadowHand} & 9/10 & 8/10 & 7/10 & 10/10 & 85.0$\%$ \\
        \hline
    \end{tabular}
    \label{tab:real}
\end{table}

To further demonstrate the robustness of our approach, we designed an \textit{extreme-case} robustness evaluation. In this setting, the robot is required to maintain a firm grasp on a target object that continuously changes its shape from (a) a water bottle to (b) a cylinder, (c) a computer mouse, (d) a Rubik's cube, and finally (e) an apple. 
% \jiayuan{WOW it runs realtime??? You should really emphasize this! And maybe, hint, in the introduction and here, this is a useful test because it is analogous to deformable objects grasping}

We conduct the robustness evaluation in the MuJoCo physics simulator \cite{todorov2012mujoco} using the ShadowHand platform.
The target object is modeled as a continuously morphing sequence of meshes that transition between distinct geometries. Each transformation takes 5 seconds and is realized through 100 intermediate interpolation waypoints. These intermediate meshes are generated offline in Blender\footnote{Blender: { \texttt{www.blender.org}}} and sequentially loaded into the simulator during runtime.
Except for the initial grasp, all subsequent grasps are initialized from the final robot configuration of the previous object. The grasp optimization algorithm runs online at 1\,Hz to update target hand poses. The robot executes these poses under the control of a PID controller operating at 200\,Hz.

Throughout the entire transformation process, the robot hand is able to sustain a stable and feasible grasp without failure.
Although such object transformations exceed typical real-world scenarios, this challenging evaluation highlights the stability of our approach under severe shape variations. The robustness of our method supports its potential applicability to important scenarios such as deformable objects grasping, tool using, human-robot interaction, etc.

\subsection{Real-Robot Experiments} \label{sec:real}

As illustrated in Fig.~\ref{fig:real-world-exp}, we conducted real-robot experiments using two setups: 1) a UFactory xArm7 robot\footnote{xArm7: \texttt{www.ufactory.cc}} equipped with the LeapHand \cite{shaw2023leaphand}, 
and 2) a UR10 robot\footnote{UR10: \texttt{www.universal-robots.com}} with the ShadowHand. 
We evaluate our methods on 4 objects, each object with 10 trials.
A trial is considered success if the grasp successfully holds the object for 3 seconds.
% For the LEAP hand experiments, a human operator moved the xArm7 to closely approximate the correct 3D grasp position of the hand relative to the object. The grasp was then executed, and the experiment was judged on the basis of whether the object was successfully grasped and held semi-firmly. 

% COMMENT ON SUCCESS RATES
% The experiments on the tomato resulted in the most successful grasps which was likely due to the simple shape and size. The tuna can, however, had the least firm. We attribute to the size being awkwardly small relative to the LEAP hand to execute the grasp. The drill had more failures due to the complex nature of the shape which resulted in grasps that knocked the drill out of position. It is interesting to note that because of the protrusions in the shape of the drill it allowed the hand to firmly grasp the drill and it resulted in the best grasps of all the objects when successful.

The results are shown in Tab.~\ref{tab:real}. On the LEAP Hand, our method achieved an average success rate of 87.5$\%$ across 40 trials, while on the ShadowHand it achieved an average success rate of 85.0$\%$. These real-robot experiments demonstrate the feasibility and effectiveness of our approach in real-world dexterous grasping scenarios.

\begin{figure}[t]
    \centering
    \includegraphics[width=0.85\linewidth]{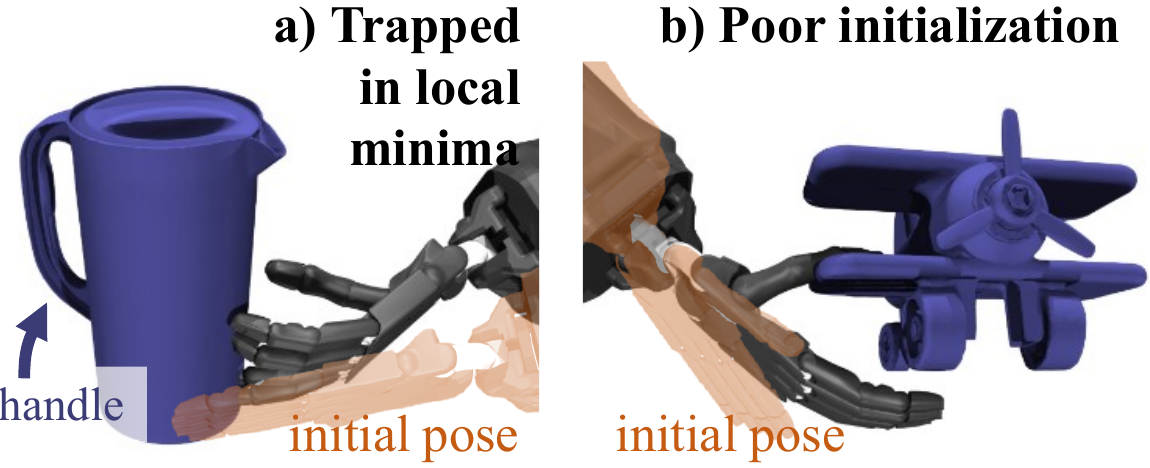}
    \caption{
    Grasp failure examples.
    }
    \label{fig:failure-case}
\end{figure}

\section{Conclusions and Future Work}
We present a grasp synthesis approach that addresses the instability and insufficient constraint of object escape in previous methods by explicitly considering possible adversarial object behaviors through a novel game-theoretic formulation. 
We formulate a two-player game where Player~1 controls the robot to generate feasible grasps and Player~2 adversarially controls the object to attempt escape.
Experimental results show that our approach can generate grasps online at 1Hz and achieves higher success rates compared to recent state-of-the-art methods by 7.51$\%$. 
Finally, real-robot experiments demonstrate the feasibility of deploying our approach on physical platforms for dexterous grasping tasks.
We believe that our game-theoretic formulation provides a promising direction for robust grasp synthesis and holds significant potential for adaptive dexterous manipulation.

Our approach suggests several future directions in both formulations and optimization algorithms. First, the proposed firm-grasp condition is defined in terms of kinematic constraints without considering friction. 
As a result, grasps generated under this condition may be overly conservative. 
Moreover, due to their size or geometry, many objects (e.g., hammers, large boxes, or cylindrical cans) are difficult to be fully constrained geometrically by the robot hand, which makes the algorithm hard to converge. 
Incorporating friction into the grasp condition represents an exciting future direction.
Second, the inner optimization problems of both players are solved using iterative local optimization. While computationally efficient, this approach suffers from being trapped in local minima. 
A failure case of pitcher grasping is illustrated in Fig.~\ref{fig:failure-case}a, where the robot hand remains trapped near its initial position and fails to reach a viable grasping region (e.g., the handle at the top).
Developing more comprehensive and theoretically grounded techniques to solve the game would be an important avenue for future work.
Third, our approach relies on \ac{PCA}-based analysis and frame alignment to initialize the relative pose between the robot and the object, which may lead to poor initializations when the object has a complex shape. 
An example of a toy airplane grasping failure is shown in Fig.~\ref{fig:failure-case}b.
Designing more adaptable initialization strategies therefore remains an important direction for improvement.

%%%%%%%%%%%%%%%%%%%%%%%%%%%%%%%%%%%%%%%%%%%%%%%%%%%%%%%%%%%%%%%%%%%%%%%%%%%%%%%%

%%%%%%%%%%%%%%%%%%%%%%%%%%%%%%%%%%%%%%%%%%%%%%%%%%%%%%%%%%%%%%%%%%%%%%%%%%%%%%%%

%%%%%%%%%%%%%%%%%%%%%%%%%%%%%%%%%%%%%%%%%%%%%%%%%%%%%%%%%%%%%%%%%%%%%%%%%%%%%%%%
% \clearpage
\bibliographystyle{IEEEtran}
\bibliography{ICRA}

\end{document}